\documentclass[runningheads]{llncs}
\usepackage{graphicx}

\usepackage{tikz}
\usepackage{comment}
\usepackage{amsmath,amssymb} 
\usepackage{color}

\usepackage[accsupp]{axessibility}  

\usepackage{wrapfig}
\usepackage{bbm}
\usepackage{booktabs}
\usepackage{makecell}
\usepackage{multirow}
\usepackage{enumerate}
\usepackage[title]{appendix}

\begin{document}
\pagestyle{headings}
\mainmatter
\def\ECCVSubNumber{5393}  

\title{Cross-Domain Cross-Set Few-Shot Learning via\\ Learning Compact and Aligned Representations} 

\titlerunning{Cross-Domain Cross-Set Few-Shot Learning}
%
\author{Wentao Chen\inst{1,2} \and
Zhang Zhang\inst{2,3} \and
Wei Wang\inst{2,3} \and
Liang Wang\inst{2,3} \and
Zilei Wang \inst{1} \and
Tieniu Tan \inst{1,2,3}}
%
\authorrunning{W. Chen et al.}
%
\institute{
University of Science and Technology of China, Hefei, China \and
Center for Research on Intelligent Perception and Computing,\\
NLPR, CASIA, Beijing, China \and
University of Chinese Academy of Sciences, Beijing, China\\
\email{wentao.chen@cripac.ia.ac.cn}, \email{zlwang@ustc.edu.cn}\\
\email{\{zzhang, wangwei, wangliang, tnt\}@nlpr.ia.ac.cn}\\
}

\maketitle
\begin{abstract}
Few-shot learning (FSL) aims to recognize novel queries with only a few support samples through leveraging prior knowledge from a base dataset. In this paper, we consider the domain shift problem in FSL and aim to address the domain gap between the support set and the query set. Different from previous cross-domain FSL work (CD-FSL) that considers the domain shift between base and novel classes, the new problem, termed cross-domain cross-set FSL (CDSC-FSL), requires few-shot learners not only to adapt to the new domain, but also to be consistent between different domains within each novel class. To this end, we propose a novel approach, namely \textit{stab}PA, to learn prototypical compact and cross-domain aligned representations, so that the domain shift and few-shot learning can be addressed simultaneously. We evaluate our approach on two new CDCS-FSL benchmarks built from the DomainNet and Office-Home datasets respectively. Remarkably, our approach outperforms multiple elaborated baselines by a large margin, e.g., improving 5-shot accuracy by 6.0 points on average on DomainNet. Code is available at \url{https://github.com/WentaoChen0813/CDCS-FSL}.

\keywords{cross-domain cross-set few-shot learning, prototypical alignment}
\end{abstract}

\section{Introduction}


Learning a new concept with a very limited number of examples is easy for human beings. However, it is quite difficult for current deep learning models, which usually require plenty of labeled data to learn generalizable and discriminative representations. To bridge the gap between humans and machines, few-shot learning (FSL) has been recently proposed \cite{vinyals2016matching,Ravi2017OptimizationAA}.

Similar to human beings, most FSL algorithms leverage prior knowledge from known classes to assist recognizing novel concepts. Typically, a FSL algorithm is composed of two phases: (i) pre-train a model on a base set that contains a large number of seen classes (called meta-training phase); (ii) transfer the pre-trained model to novel classes with a small labeled support set and test it with a query set (meta-testing phase). 
Despite great progresses on FSL algorithms \cite{finn2017model,rusu2018metalearning,vinyals2016matching,tian2020rethinking}, most previous studies adopt a single domain assumption, where all images in both meta-training and meta-testing phases are from a single domain. Such assumption, however, may be easily broken in real-world applications. Considering a concrete example of online shopping, a clothing retailer commonly shows several high-quality pictures taken by photographers for each fashion product (support set), while customers may use their cellphone photos (query set) to match the displayed pictures of their expected products. In such case, there is a distinct domain gap between the support set and the query set. Similar example can be found in security surveillance: given the low-quality picture of a suspect captured at night (query set), the surveillance system is highly expected to recognize its identity based on a few high-quality registered photos (e.g., ID card). With such domain gap, FSL models will face more challenges besides limited support data.

\begin{figure*}[t]
    \centering
    \includegraphics[width=.95\linewidth]{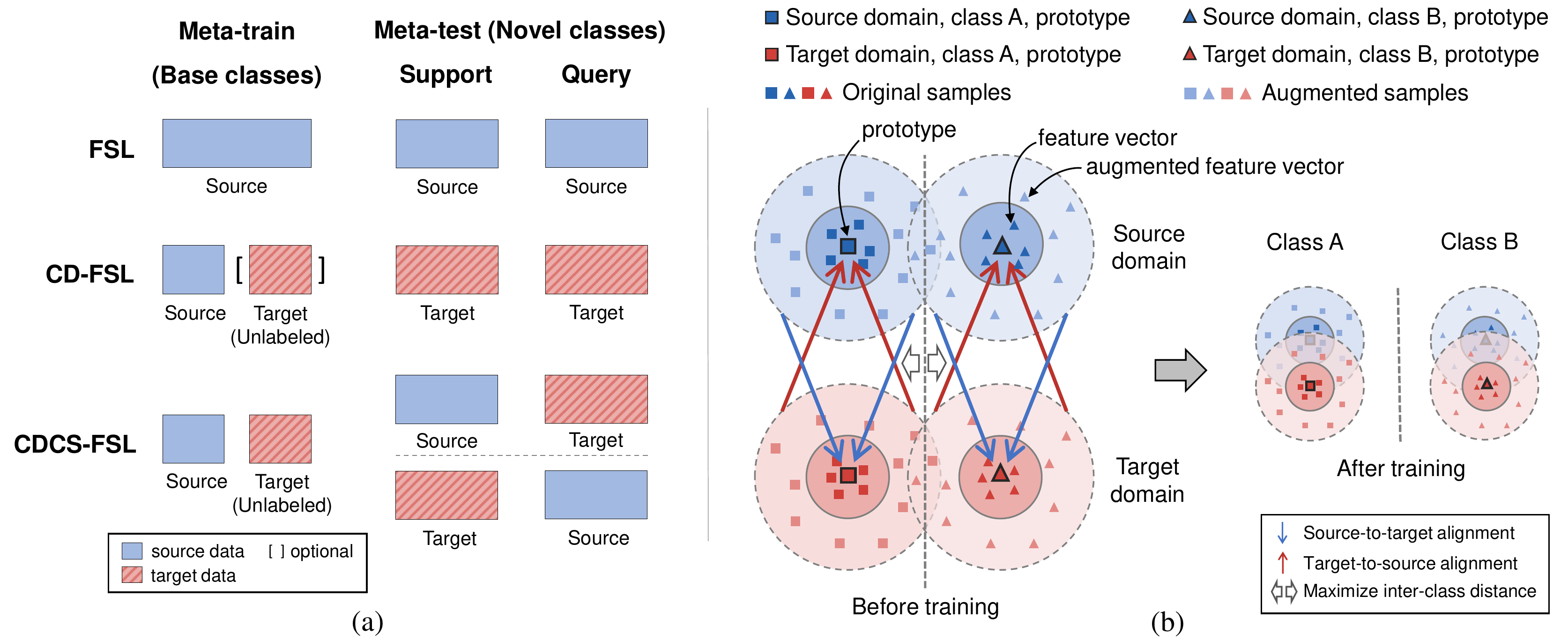}
    \caption{\textbf{Problem setup and motivation.} (a) CD-FSL considers the domain shift between the meta-training and meta-testing phases, while CDCS-FSL considers the domain shift between the support set and the query set in the meta-testing phase. Following previous CD-FSL work \cite{phoo2021selftraining}, unlabeled target domain data are used in CDCS-FSL to assistant model training. (b) We propose a bi-directional prototypical alignment framework to address CDCS-FSL, which pushes feature vectors of one domain to be gathered around the corresponding prototype in the other domain bi-directionally, and separates feature vectors from different classes.}
    \label{fig:setting}
\end{figure*}

In this paper, we consider the above problem in FSL and propose a new setting to address the domain gap between the support set and the query set. Following previous FSL work, a large base set from the source domain is available for meta-training. Differently, during meta-testing, only the support set or the query set is from the source domain, while the other is from a different target domain. Some recent studies also consider the cross-domain few-shot learning problem (CD-FSL) \cite{guo2020broader,Tseng2020Cross-Domain,phoo2021selftraining}. However, the domain shift in CD-FSL occurs between the meta-training and meta-testing phases. In other words, both the support and query sets in the meta-testing phase are still from the same domain (pictorial illustration is given in Figure \ref{fig:setting} (a)). To distinguish the considered setting from CD-FSL, we name this setting as cross-domain cross-set few-shot learning (CDCS-FSL), as the support set and the query set are across different domains. Compared to CD-FSL, the domain gap within each novel class imposes more requirements to learn a well-aligned feature space. Nevertheless, in terms of the above setting, it is nearly intractable to conquer the domain shift due to the very limited samples of the target domain, e.g., the target domain may contain only one support (or query) image. Thus, we follow the CD-FSL literature \cite{phoo2021selftraining} to use unlabeled auxiliary data from the target domain to assist model training. Note that we do not suppose that the auxiliary data are from novel classes. Therefore, we can collect these data from some common-seen classes (e.g., base classes) without any annotation costs. 

One may notice that re-collecting a few support samples from the same domain as the query set can `simply' eliminate the domain gap. However, it may be intractable to re-collect support samples in some real few-shot applications, e.g., re-collecting ID photos for all persons is difficult. Besides, users sometimes not only want to get the class labels, but more importantly they'd like to retrieve the support images themselves (like the high-quality fashion pictures). Therefore, the CDCS-FSL setting can not be simply transferred into previous FSL and CD-FSL settings.

To address the CDCS-FSL problem, we propose a simple but effective bi-directional prototypical alignment framework to learn compact and cross-domain aligned representations, which is illustrated in Figure \ref{fig:setting} (b). The main idea of our approach is derived from two intuitive insights: (i) we need aligned representations to alleviate the domain shift between the source and target domains, and (ii) compact representations are desirable to learn a center-clustered class space, so that a small support set can better represent a new class. Specifically, given the labeled base set in the source domain and the unlabeled auxiliary set in the target domain, we first assign pseudo labels to the unlabeled data considering that pseudo labels can preserve the coarse semantic similarity with the visual concepts in source domain. Then, we minimize the point-to-set distance between the prototype (class center) in one domain and the corresponding feature vectors in the other domain bi-directionally. As results, the feature vectors of the source (or target) domain will be gathered around the prototype in the other domain, thus reducing the domain gap and intra-class variance simultaneously. Moreover, the inter-class distances are maximized to attain a more separable feature space. Furthermore, inspired by the fact that data augmentation even with strong transformations generally does not change sample semantics, we augment samples in each domain, and suppose that the augmented samples between different domains should also be aligned. Since these augmented samples enrich the data diversity, they can further encourage to learn the underlying invariance and strengthen the cross-domain alignment.

Totally, we summarize all the above steps into one approach termed ``\textbf{St}rongly \textbf{A}ugmented \textbf{B}i-directional \textbf{P}rototypical \textbf{A}lignment", or \textit{stab}PA. We evaluate its effectiveness on two new CDCS-FSL benchmarks built from the DomainNet \cite{peng2019moment} and Office-Home \cite{venkateswara2017deep} datasets. Remarkably, the proposed \textit{stab}PA achieves the best performance over both benchmarks and outperforms other baselines with a large margin, e.g., improving 5-shot accuracy by 6.0 points on average on the DomainNet dataset.

In summary, our contributions are three-fold: 
\begin{itemize}
    \item We consider a new FSL setting, CDCS-FSL, where a domain gap exists between the support set and the query set.
    \item We propose a novel approach, namely \textit{stab}PA, to address the CDCS-FSL problem, of which the key is to learn prototypical compact and domain aligned representations.
    \item Extensive experiments demonstrate that \textit{stab}PA can learn discriminative and generalizable representations and outperforms all baselines by a large margin on two CDCS-FSL benchmarks.
\end{itemize}

\section{Related Work}

FSL aims to learn new classes with very few labeled examples. Most studies follow a meta-learning paradigm \cite{vilalta2002perspective}, where a meta-learner is trained on a series of training tasks (episodes) so as to enable fast adaptation to new tasks. The meta-learner can take various forms, such as an LSTM network \cite{Ravi2017OptimizationAA}, initial network parameters \cite{finn2017model}, or closed-form solvers \cite{rusu2018metalearning}. Recent advances in pre-training techniques spawn another FSL paradigm. In \cite{chen2018a}, the authors show that a simple pre-training and fine-tuning baseline can achieve competitive performance with respect to the SOTA FSL models. In \cite{tian2020rethinking,chen2021few}, self-supervised pre-training techniques have proven to be useful for FSL. Our approach also follows the pre-training paradigm, and we further expect the learned representations to be compact and cross-domain aligned to address the CDCS-FSL problem.

CD-FSL \cite{guo2020broader,Tseng2020Cross-Domain,phoo2021selftraining,ijcai2021-149,liang2021boosting,guan2020large,fu2021meta} considers the domain shift problem between the base classes and the novel classes. Due to such domain gap, \cite{chen2018a} show that meta-learning approaches fail to adapt to novel classes. To alleviate this problem, \cite{Tseng2020Cross-Domain} propose a feature-wise transformation layer to learn rich representations that can generalize better to other domains. However, they need to access multiple labeled data sources with extra data collection costs. \cite{phoo2021selftraining} solve this problem by exploiting additional unlabeled target data with self-supervised pre-training techniques. Alternatively, \cite{guan2020large} propose to utilize the semantic information of class labels to minimize the distance between source and target domains. Without the need for extra data or language annotations, \cite{ijcai2021-149} augment training tasks in an adversarial way to improve the generalization capability. 

Using target domain images to alleviate domain shift is related to the field of domain adaptation (DA). Early efforts align the marginal distribution of each domain by minimizing a pre-defined discrepancy, such as $\mathcal{H}\Delta \mathcal{H}$-divergence \cite{ben2010theory} or Maximum Mean Discrepancy (MMD) \cite{gretton2006kernel}. Recently, adversarial-based methods adopt a discriminator \cite{goodfellow2014generative} to approximate the domain discrepancy, and learn domain-invariant distribution at image level \cite{hoffman2018cycada}, feature level \cite{NEURIPS2018_ab88b157} or output level \cite{tsai2018learning}. Another line of studies assign pseudo labels to unlabeled target data \cite{zheng2021rectifying,zhang2021prototypical,NEURIPS2019_6da9003b}, and directly align the feature distribution within each class. Although these DA methods are related to our work, they usually assume that the testing stage shares the same class space as the training stage, which is broken by the setting of FSL. Open-set DA \cite{panareda2017open,saito2018open} and Universal DA \cite{saito2020universal,you2019universal} consider the existence of unseen classes. However, they merely mark them as `unknown'. In this work, we are more interested in addressing the domain shift for these unseen novel classes within a FSL assumption.

\section{Problem Setup}

\label{setup}
Formally, a FSL task often adopts a setting of $\mathrm{N}$-way-$\mathrm{K}$-shot classification, which aims to discriminate between $\mathrm{N}$ novel classes with $\mathrm{K}$ exemplars per class. Given a support set $\mathcal{S} = \{(x_i, y_i)\}_{i=1}^{N\times K}$  where $x_i \in \mathcal{X_N} $ denotes a data sample in novel classes and $y_i\in Y_\mathcal{N}$ is the class label, the goal of FSL is to learn a mapping function $\phi: \phi(x_q)\rightarrow y_q $ which classifies a query sample $x_q$ in the query set $\mathcal{Q}$ to the class label $y_q\in Y_\mathcal{N}$.   Besides $\mathcal{S}$ and $\mathcal{Q}$, a large labeled dataset $\mathcal{B} \subset \mathcal{X_B} \times \mathcal{Y_B}$ (termed base set) is often provided for meta-training, where $\mathcal{X_B} $ and $\mathcal{Y_B}$ do not overlap with $\mathcal{X_N}$  and $\mathcal{Y_N}$.

Conventional FSL studies assume the three sets $\mathcal{S}$, $\mathcal{Q}$ and $\mathcal{B}$ are from the same domain. In this paper, we consider the domain gap between the support set and the query set (only one is from the same domain as the base set, namely the source domain $\mathcal{D}_s$, and the other is from a new target domain $\mathcal{D}_t$). Specifically, this setting has two situations: 
\begin{enumerate}[(i)]
    \item $\mathcal{D}_s-\mathcal{D}_t$: the support set is from the source domain and the query set is from the target domain, i.e., $\mathcal{S}\subset \mathcal{D}_s$ and $\mathcal{Q}\subset \mathcal{D}_t$.
    \item $\mathcal{D}_t-\mathcal{D}_s$: the support set is from the target domain and the query set is from the source domain, i.e., $\mathcal{S}\subset \mathcal{D}_t$ and $\mathcal{Q}\subset \mathcal{D}_s$.
\end{enumerate}
As the support set and the query set are across different domains, we name this setting as cross-domain cross-set few-shot learning (CDCS-FSL). Besides the above three sets, to facilitate crossing the domain gap, an unlabeled auxiliary set $\mathcal{U}$ from the target domain is available in the meta-training phase, where the data from novel classes are manually removed to promise they are not seen in meta-training.

\section{Approach}

\begin{figure*}[t]
    \centering
    \includegraphics[width=\linewidth]{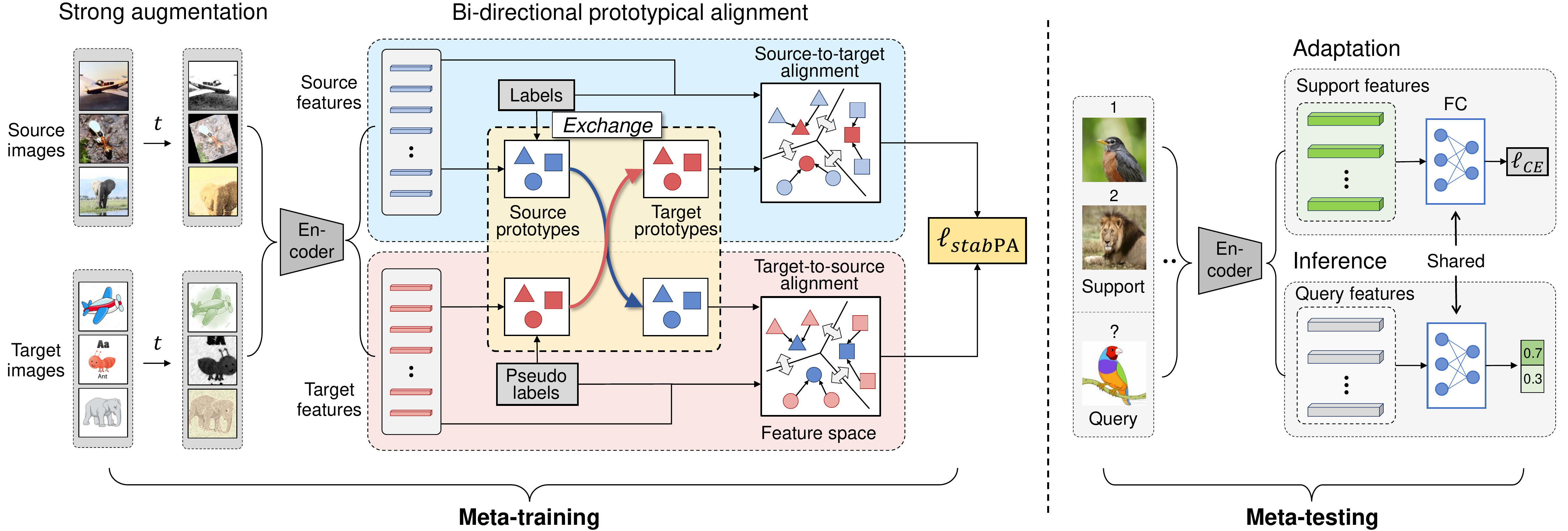}
    \caption{\textbf{Framework.} In the meta-training stage, we train a feature extractor within the bi-directional prototypical alignment framework to learn compact and aligned representations. In the meta-testing stage, we fix the feature extractor and train a new classification head with the support set, and then evaluate the model on the query set.}
    \label{fig:framework}
\end{figure*}

Briefly, our approach contains two stages: 1) In the meta-training stage, we train a feature extractor $f: x_i \rightarrow f(x_i)$ with the base set $\mathcal{B}$ and the unlabeled auxiliary set $\mathcal{U}$; 2) In the meta-testing stage, we fix the feature extractor and train a linear classification head $g: f(x_i)\rightarrow y_i$ on the support set $\mathcal{S}$, and the entire model $\phi = g\circ f$ is used to predict the labels for the query set $\mathcal{Q}$. The framework of our approach is illustrated in Figure \ref{fig:framework}.



\subsection{Bi-directional Prototypical Alignment}



A straightforward way to align feature distributions is through estimating class centers (prototypes) in both source and target domains. With the labeled base data, it is easy to estimate prototypes for the source domain. However, it is difficult to estimate prototypes in the target domain with only unlabeled data available. To address this issue, we propose to assign pseudo labels to the unlabeled data and then use the pseudo labels to approximate prototypes. The insight is that the pseudo labels can preserve the coarse semantic similarity even under domain or category shift (e.g., a painting tiger could be more likely to be pseudo-labeled as a cat rather than a tree). Aggregating samples with the same pseudo label can extract the shared semantics across different domains.

Specifically, given the source domain base set $\mathcal{B}$ and the target domain unlabeled set $\mathcal{U}$, we first assign pseudo labels to the unlabeled samples with an initial classifier $\phi_0$ trained on the base set and obtain $\mathcal{\hat{U}} = \{(x_i, \hat{y}_i)|x_i\in \mathcal{U}\}$, where $\hat{y}_i =\phi_0 (x_i )$ is the pseudo label. Then, we obtain the source prototypes $\{p_k^s\}_{k=1}^{|\mathcal{Y_B}|}$ and the target prototypes $\{p_k^t\}_{k=1}^{|\mathcal{Y_B}|}$ by averaging the feature vectors with the same label (or pseudo label). It should be noted that the prototypes are estimated on the entire datasets $\mathcal{B}$ and $\mathcal{\hat{U}}$, and adjusted together with the update of the feature extractor and pseudo labels (details can be found below).

With the obtained prototypes, directly minimizing the \textit{point-to-point} distance between two prototypes $p_k^s$ and $p_k^t$ can easily reduce the domain gap for the class $k$. However, this may make the feature distribution of different classes mix together and the discrimination capability of the learned representations is still insufficient. To overcome these drawbacks, we propose to minimize the \textit{point-to-set} distance across domains in a \textit{bi-directional} way. That is, we minimize the distance between the \textit{prototype in one domain} and the corresponding \textit{feature vectors in the other domain}, and meanwhile maximize the feature distance between different classes. In this way, we can not only align features across domains, but also simultaneously obtain compact feature distributions for both domains to suit the requirement of few-shot learning.

Concretely, for a source sample $(x_i^s, y_i^s)\in \mathcal{B}$ of the $q$-th class (i.e., $y_i^s=q$), we minimize its feature distance to the prototype $p_q^t$ in the target domain, and meanwhile maximize its distances to the prototypes of other classes. Here, a softmax loss function for the source-to-target alignment is formulated as:
\begin{equation}
    \ell_{s-t}(x_i^s, y_i^s) = -\log \frac{\exp{(-||f(x_i^s) - p_q^t||^2/ \tau) }}{\sum_{k=1}^{|\mathcal{Y_B}|}\exp{(-||f(x_i^s) - p_k^t||^2/ \tau) }},
\end{equation}
where $\tau$ is a temperature factor. To get a better feature space for the target domain, a similar target-to-source alignment loss is applied for each target sample $(x_i^t, \hat{y}_i^t)\in \mathcal{\hat{U}}$  with $\hat{y}_i^t = q$:
\begin{equation}
    \ell_{t-s}(x_i^t, \hat{y}_i^t) = -\log \frac{\exp{(-||f(x_i^t) - p_q^s||^2 / \tau)}}{\sum_{k=1}^{|\mathcal{Y_B}|}\exp{(-||f(x_i^t) - p_k^s||^2/ \tau)}}.
\end{equation}
Since the initial pseudo labels are more likely to be incorrect, we gradually increase the weights of these two losses following the principle of curriculum learning \cite{bengio2009curriculum}. For the source-to-target alignment, the loss weight starts from zero and converges to one, formulated as:
\begin{equation}
    w(t) = \frac{2}{1+\exp{( -t / T_{max})}} - 1, 
\end{equation}
where $t$ is the current training step and $T_{max}$ is the maximum training step. For the target-to-source alignment, since the pseudo labels become more confident along with the training process, a natural curriculum is achieved by setting a confidence threshold to filter out the target samples with low confidence pseudo labels \cite{sohn2020fixmatch}. 

Together, the total loss for the bi-directional prototypical alignment is
\begin{equation}
\ell_{b\text{PA}} = 
    \frac{1}{|\mathcal{B}|} \sum_{i=1}^{|\mathcal{B}|} w(t) \ell_{s-t}(x_i^s, y_i^s)
    + \frac{1}{|\mathcal{\hat{U}}|} \sum_{i=1}^{|\mathcal{\hat{U}}|} \mathbbm{1}(p(\hat{y}_i^t) > \beta) \ell_{t-s}(x_i^t, \hat{y}_i^t), 
\end{equation}
where $p(\cdot)$ is the confidence of a pseudo label, and $\beta$ is the confidence threshold below which the data samples will be dropped. 

\textbf{Updating Pseudo Label.}
The pseudo labels are initially predicted by a classifier $\phi_0$ pre-trained on the base set $\mathcal{B}$. As the representations are updated, we update the pseudo labels by re-training a classifier $\phi_t = h\circ f$ based on the current feature extractor $f$, where $h$ is a linear classification head for the base classes. The final pseudo labels are updated by linear interpolation between the predictions of the initial classifier $\phi_0$ and the online updated classifier $\phi_t$:
\begin{equation}
    \hat{y}_i = \arg\max_{k\in \mathcal{Y_B}} \, \lambda \phi_0(k|x_i) + (1-\lambda) \phi_t(k|x_i), 
\end{equation}
where $\lambda$ is the interpolation coefficient. The combination of these two classifiers makes it possible to rectify the label noise of the initial classifier, and meanwhile inhibit the rapid change of pseudo labels of online classifier especially in the early training stage. 

\textbf{Generating Prototypes.}
\label{Prototype}
Note that we are intended to estimate the prototypes on the entire dataset and update them with representation learning. For the source domain, instead of calculating the mean value of intra-class samples in the feature space, a cheaper way is to approximate prototypes with the normalized weights of the classification head $h$, as the classifier weights tend to align with class centers in order to reduce classification errors \cite{qiao2018few}. Specifically, we set the source prototypes as $p_k^s = W_k$, where $W_k$ is the normalized classification weight for the $k$-th class. For the target domain, we adopt the momentum technique to update prototypes. The prototypes are initialized as zeros. At each training step, we first estimate the prototypes using target samples in current batch with their pseudo labels. Then, we update the target prototype $p_k^t$ as:
\begin{equation}
    p_k^t \longleftarrow m p_k^t + (1-m) \frac{1}{n_k} \sum_{i=1}^{|\mathcal{\hat{U}}_b|} \mathbbm{1}(\hat{y}_i^t=k)f(x_i^t), 
\end{equation}
where $n_{k}$ is the number of the target samples classified into the $k$-th class in a target batch $\mathcal{\hat{U}}_b$, and $m$ is the momentum term controlling the update speed.

\subsection{stabPA}

Strong data augmentation has proved to be effective for learning generalizable representations, especially in self-supervised representation learning studies \cite{he2020momentum,chen2020simple}. Given a sample $x$, strong data augmentation generates additional data points $\{\widetilde{x}_i\}_{i=1}^{n}$  by applying various intensive image transformations. The assumption behind strong data augmentation is that the image transformations do not change the semantics of original samples. 

In this work, we further hypothesize that strongly augmented intra-class samples in different domains can also be aligned. It is expected that strong data augmentation can further strengthen the learning of cross-domain representations, since stronger augmentation provides more diverse data samples and makes the learned aligned representations more robust for various transformations in both the source and target domains.

Following this idea, we extend the bi-directional prototypical alignment with strong data augmentation and the entire framework is termed \textit{stab}PA. Specifically, for a source sample $(x_i^s, y_i^s)$ and a target sample $(x_i^t, \hat{y}_i^t)$, we generate their augmented versions $(\widetilde{x}_i^s, y_i^s)$ and $(\widetilde{x}_i^t, \hat{y}_i^t)$. Within the bi-directional prototypical alignment framework, we minimize the feature distance of a strongly augmented image to its corresponding prototype in the other domain, and maximize its distances to the prototypes of other classes. Totally, the \textit{stab}PA loss is
\begin{equation}
\ell_{stab\text{PA}} = 
    \frac{1}{|\mathcal{\widetilde{B}}|} \sum_{i=1}^{|\mathcal{\widetilde{B}}|} w(t) \ell_{s-t}(\widetilde{x}_i^s, y_i^s)
    + \frac{1}{|\mathcal{\widetilde{U}}|} \sum_{i=1}^{|\mathcal{\widetilde{U}}|} \mathbbm{1}(p(\hat{y}_i^t) > \beta) \ell_{t-s}(\widetilde{x}_i^t, \hat{y}_i^t),
\end{equation}
where $\mathcal{\widetilde{B}}$ and $\mathcal{\widetilde{U}}$ are the augmented base set and unlabeled auxiliary  set, respectively.

To perform strong data augmentation, we apply random crop, Cutout \cite{devries2017improved}, and RandAugment \cite{cubuk2020randaugment}. RandAugment comprises 14 different transformations and randomly selects a fraction of transformations for each sample. In our implementation, the magnitude for each transformation is also randomly selected, which is similar to \cite{sohn2020fixmatch}.

\section{Experiments}

\subsection{Datasets}

\quad\ \textbf{DomainNet.} DomainNet \cite{peng2019moment} is a large-scale multi-domain image dataset. It contains 345 classes in 6 different domains. In experiments, we choose the \textit{real} domain as the source domain and choose one domain from \textit{painting}, \textit{clipart} and \textit{sketch} as the target domain. We randomly split the classes into 3 parts: base set (228 classes), validation set (33 classes) and novel set (65 classes), and discard 19 classes with too few samples. 

\textbf{Office-Home.} Office-Home \cite{venkateswara2017deep} contains 65 object classes usually found in office and home settings. We randomly select 40 classes as the base set, 10 classes as the validation set, and 15 classes as the novel set. There are 4 domains for each class: \textit{real}, \textit{product}, \textit{clipart} and \textit{art}. We set the source domain as \textit{real} and choose the target domain from the other three domains.

In both datasets, we construct the unlabeled auxiliary set by collecting data from the base and validation sets of the target domain and removing their labels. These unlabeled data combined with the labeled base set are used for meta-training. The validation sets in both domains are used to tune hyper-parameters. Reported results are averaged across 600 test episodes from the novel set.

\subsection{Comparison Results}

\begin{table*}[t!]
    \small
    \centering
    \caption{Comparison to baselines on the DomainNet dataset. We denote `r' as \textit{real}, `p' as \textit{painting}, `c' as \textit{clipart} and `s' as \textit{sketch}. We report 5-way 1-shot and 5-way 5-shot accuracies with 95\% confidence interval.}
    \resizebox{.95\linewidth}{!}{
    \begin{tabular}{lccccccc}
    \toprule
    & \multicolumn{7}{c}{5-way 1-shot} \\
    \cmidrule(lr){2-8}
    Method & r-r & r-p & p-r & r-c & c-r & r-s & s-r \\
    \midrule
    ProtoNet \cite{NIPS2017_cb8da676} & 63.43\tiny$\pm$0.90 & 45.36\tiny$\pm$0.81 & 45.25\tiny$\pm$0.97 & 44.65\tiny$\pm$0.81 & 47.50\tiny$\pm$0.95 & 39.28\tiny$\pm$0.77 & 42.85\tiny$\pm$0.89 \\
    RelationNet \cite{sung2018learning} & 59.49\tiny$\pm$0.91 & 42.69\tiny$\pm$0.77 & 43.04\tiny$\pm$0.97 & 44.12\tiny$\pm$0.81 & 45.86\tiny$\pm$0.95 & 36.52\tiny$\pm$0.73 & 41.29\tiny$\pm$0.96 \\
    MetaOptNet \cite{lee2019meta} & 61.12\tiny$\pm$0.89 & 44.02\tiny$\pm$0.77 & 44.31\tiny$\pm$0.94 & 42.46\tiny$\pm$0.80 & 46.15\tiny$\pm$0.98 & 36.37\tiny$\pm$0.72 & 40.27\tiny$\pm$0.95 \\
    Tian et al. \cite{tian2020rethinking} & 67.18\tiny$\pm$0.87 & 46.69\tiny$\pm$0.86 & 46.57\tiny$\pm$0.99 & 48.30\tiny$\pm$0.85 & 49.66\tiny$\pm$0.98 & 40.23\tiny$\pm$0.73 & 41.90\tiny$\pm$0.86 \\
    DeepEMD \cite{Zhang_2020_CVPR} & 67.15\tiny$\pm$0.87 & 47.60\tiny$\pm$0.87 & 47.86\tiny$\pm$1.04 & 49.02\tiny$\pm$0.83 & 50.89\tiny$\pm$1.00 & 42.75\tiny$\pm$0.79 & 46.02\tiny$\pm$0.93 \\
    ProtoNet+FWT \cite{Tseng2020Cross-Domain} & 62.38\tiny$\pm$0.89 & 44.40\tiny$\pm$0.80 & 45.32\tiny$\pm$0.97 & 43.95\tiny$\pm$0.80 & 46.32\tiny$\pm$0.92 & 39.28\tiny$\pm$0.74 & 42.18\tiny$\pm$0.95\\
    ProtoNet+ATA \cite{ijcai2021-149} & 61.97\tiny$\pm$0.87 & 45.59\tiny$\pm$0.84 & 45.90\tiny$\pm$0.94 & 44.28\tiny$\pm$0.83 & 47.69\tiny$\pm$0.90 & 39.87\tiny$\pm$0.81 & 43.64\tiny$\pm$0.95 \\
    S2M2 \cite{mangla2020charting} & 67.07\tiny$\pm$0.84 & 46.84\tiny$\pm$0.82 & 47.03\tiny$\pm$0.95 & 47.75\tiny$\pm$0.83 & 48.27\tiny$\pm$0.91 & 39.78\tiny$\pm$0.76 & 40.11\tiny$\pm$0.91 \\
    Meta-Baseline \cite{chen2021meta} & \textbf{69.46\tiny$\pm$0.91} & \textbf{48.76\tiny$\pm$0.85} & 48.90\tiny$\pm$1.12 & \textbf{49.96\tiny$\pm$0.85} & \textbf{52.67\tiny$\pm$1.08} & \textbf{43.08\tiny$\pm$0.80} & \textbf{46.22\tiny$\pm$1.04}\\
    $stab\text{PA}^-$ \scriptsize(\textit{Ours}) & 68.48\tiny$\pm$0.87 & 48.65\tiny$\pm$0.89 & \textbf{49.14\tiny$\pm$0.88} & 45.86\tiny$\pm$0.85 & 48.31\tiny$\pm$0.92 & 41.74\tiny$\pm$0.78 & 42.17\tiny$\pm$0.95 \\
    \cmidrule(lr){2-8}
    DANN \cite{ganin2016domain} & - & 45.94\tiny$\pm$0.84 & 46.85\tiny$\pm$0.97 & 47.31\tiny$\pm$0.86 & 50.02\tiny$\pm$0.94 & 42.44\tiny$\pm$0.79 & 43.66\tiny$\pm$0.92 \\
    PCT \cite{tanwisuth2021a} & -  & 47.14\tiny$\pm$0.89 & 47.31\tiny$\pm$1.04 & \textbf{50.04\tiny$\pm$0.85} & 49.83\tiny$\pm$0.98 & 39.10\tiny$\pm$0.76 & 39.92\tiny$\pm$0.95 \\
    Mean Teacher \cite{NIPS2017_68053af2} & -  & 46.92\tiny$\pm$0.83 & 46.84\tiny$\pm$0.96 & 48.48\tiny$\pm$0.81 & 49.60\tiny$\pm$0.97 & 43.39\tiny$\pm$0.81 & 44.52\tiny$\pm$0.89 \\
    FixMatch \cite{sohn2020fixmatch} & - & \textbf{48.86\tiny$\pm$0.87} & \textbf{49.15\tiny$\pm$0.93} & 48.70\tiny$\pm$0.82 & 49.18\tiny$\pm$0.93 & \textbf{44.48\tiny$\pm$0.80} & \textbf{45.97\tiny$\pm$0.95} \\
    STARTUP \cite{phoo2021selftraining} & - & 47.53\tiny$\pm$0.88 & 47.58\tiny$\pm$0.98 & 49.24\tiny$\pm$0.87 & \textbf{51.32\tiny$\pm$0.98} & 43.78\tiny$\pm$0.82 & 45.23\tiny$\pm$0.96 \\
    DDN \cite{islam2021dynamic} & - & 48.83\tiny$\pm$0.84 & 48.11\tiny$\pm$0.91 & 48.25\tiny$\pm$0.83 & 48.46\tiny$\pm$0.93 & 43.60\tiny$\pm$0.79 & 43.99\tiny$\pm$0.91\\
    \cmidrule(lr){3-8}
    \textit{stab}PA \scriptsize(\textit{Ours}) & - & \textbf{53.86\tiny$\pm$0.89} & \textbf{54.44\tiny$\pm$1.00} & \textbf{56.12\tiny$\pm$0.83} & \textbf{56.57\tiny$\pm$1.02} & \textbf{50.85\tiny$\pm$0.86} & \textbf{51.71\tiny$\pm$1.01} \\
    
    \midrule
    & \multicolumn{7}{c}{5-way 5-shot}\\
    \cmidrule(lr){2-8}
    ProtoNet \cite{NIPS2017_cb8da676} & 82.79\tiny$\pm$0.58 & 57.23\tiny$\pm$0.79 & 65.60\tiny$\pm$0.95 & 58.04\tiny$\pm$0.81 & 65.91\tiny$\pm$0.78 & \textbf{51.68\tiny$\pm$0.81} & 59.46\tiny$\pm$0.85 \\
    RelationNet \cite{sung2018learning} & 77.68\tiny$\pm$0.62 & 52.63\tiny$\pm$0.74 & 61.18\tiny$\pm$0.90 & 57.24\tiny$\pm$0.80 & 62.65\tiny$\pm$0.81 & 47.32\tiny$\pm$0.75 & 56.39\tiny$\pm$0.88 \\
    MetaOptNet \cite{lee2019meta} & 80.93\tiny$\pm$0.60 & 56.34\tiny$\pm$0.76 & 63.20\tiny$\pm$0.89 & 57.92\tiny$\pm$0.79 & 63.51\tiny$\pm$0.82 & 48.20\tiny$\pm$0.79 & 55.65\tiny$\pm$0.85 \\
    Tian et al. \cite{tian2020rethinking} & 84.50\tiny$\pm$0.55 & 56.87\tiny$\pm$0.84 & 63.90\tiny$\pm$0.95 & 59.67\tiny$\pm$0.84 & 65.33\tiny$\pm$0.80 & 50.41\tiny$\pm$0.80 & 56.95\tiny$\pm$0.84 \\
    DeepEMD \cite{Zhang_2020_CVPR} & 82.79\tiny$\pm$0.56 & 56.62\tiny$\pm$0.78 & 63.86\tiny$\pm$0.93 & 60.43\tiny$\pm$0.82 & 67.46\tiny$\pm$0.78 & 51.66\tiny$\pm$0.80 & 60.39\tiny$\pm$0.87 \\
    ProtoNet+FWT \cite{Tseng2020Cross-Domain} & 82.42\tiny$\pm$0.55 & 57.18\tiny$\pm$0.77 & 65.64\tiny$\pm$0.93 & 57.42\tiny$\pm$0.77 & 65.11\tiny$\pm$0.83 & 50.69\tiny$\pm$0.77 & 59.58\tiny$\pm$0.84 \\
    ProtoNet+ATA \cite{ijcai2021-149} & 81.96\tiny$\pm$0.57 & 57.69\tiny$\pm$0.83 & 64.96\tiny$\pm$0.93 & 56.90\tiny$\pm$0.84 & 64.08\tiny$\pm$0.86 & 51.67\tiny$\pm$0.80 & 60.78\tiny$\pm$0.86 \\
    S2M2 \cite{mangla2020charting} & 85.79\tiny$\pm$0.52 & 58.79\tiny$\pm$0.81 & 65.67\tiny$\pm$0.90 & \textbf{60.63\tiny$\pm$0.83} & 63.57\tiny$\pm$0.88 & 49.43\tiny$\pm$0.79 & 54.45\tiny$\pm$0.89 \\
    Meta-Baseline \cite{chen2021meta} & 83.74\tiny$\pm$0.58 & 56.07\tiny$\pm$0.79 & 65.70\tiny$\pm$0.99 & 58.84\tiny$\pm$0.80 & \textbf{67.89\tiny$\pm$0.91} & 50.27\tiny$\pm$0.76 & \textbf{61.88\tiny$\pm$0.94} \\
    $stab\text{PA}^-$ \scriptsize(\textit{Ours}) & \textbf{85.98\tiny$\pm$0.51} & \textbf{59.92\tiny$\pm$0.85} & \textbf{67.10\tiny$\pm$0.93} & 57.10\tiny$\pm$0.88 & 62.90\tiny$\pm$0.83 & 51.03\tiny$\pm$0.85 & 57.11\tiny$\pm$0.93\\
    \cmidrule(lr){2-8}
    DANN \cite{ganin2016domain} & - & 56.83\tiny$\pm$0.86 & 64.29\tiny$\pm$0.94 & 59.42\tiny$\pm$0.84 & 66.87\tiny$\pm$0.78 & 53.47\tiny$\pm$0.75 & 60.14\tiny$\pm$0.81 \\
    PCT \cite{tanwisuth2021a} & - & 56.38\tiny$\pm$0.87 & 64.03\tiny$\pm$0.99 & 61.15\tiny$\pm$0.80 & 66.19\tiny$\pm$0.82 & 46.77\tiny$\pm$0.74 & 53.91\tiny$\pm$0.90 \\
    Mean Teacher \cite{NIPS2017_68053af2} & - & 57.74\tiny$\pm$0.84 & 64.97\tiny$\pm$0.94 & 61.54\tiny$\pm$0.84 & 67.39\tiny$\pm$0.89 & 54.57\tiny$\pm$0.79 & 60.04\tiny$\pm$0.86 \\
    FixMatch \cite{sohn2020fixmatch} & - & 61.62\tiny$\pm$0.79 & 67.46\tiny$\pm$0.89 & \textbf{61.94\tiny$\pm$0.82} & 66.72\tiny$\pm$0.81 & \textbf{55.26\tiny$\pm$0.83} & \textbf{62.46\tiny$\pm$0.87} \\
    STARTUP \cite{phoo2021selftraining} & - & 58.13\tiny$\pm$0.82 & 65.27\tiny$\pm$0.92 & 61.51\tiny$\pm$0.86 & \textbf{67.95\tiny$\pm$0.78} & 54.89\tiny$\pm$0.81 & 61.97\tiny$\pm$0.88 \\
    DDN \cite{islam2021dynamic} & - & \textbf{61.98\tiny$\pm$0.82} & \textbf{67.69\tiny$\pm$0.88} & 61.07\tiny$\pm$0.84 & 65.58\tiny$\pm$0.79 & 54.35\tiny$\pm$0.83 & 60.37\tiny$\pm$0.88 \\
    \cmidrule(lr){3-8}
    \textit{stab}PA \scriptsize(\textit{Ours}) & - & \textbf{65.65\tiny$\pm$0.74} & \textbf{73.63\tiny$\pm$0.82} & \textbf{67.32\tiny$\pm$0.80} & \textbf{74.41\tiny$\pm$0.76} & \textbf{61.37\tiny$\pm$0.82} & \textbf{68.93\tiny$\pm$0.87} \\
    \bottomrule
    \end{tabular}
    }
    \label{tab:domainnet}
\end{table*}

\begin{table*}[h!]
    \small
    \centering
    \caption{Comparison results on Office-Home. We denote `r' as \textit{real}, `p' as \textit{product}, `c' as \textit{clipart} and `a' as \textit{art}. Accuracies are reported with 95\% confidence intervals.}
    \resizebox{.95\linewidth}{!}{
    \begin{tabular}{lccccccc}
    \toprule
    & \multicolumn{7}{c}{5-way 1-shot} \\
    \cmidrule(lr){2-8}
    Method & r-r & r-p & p-r & r-c & c-r & r-a & a-r \\
    \midrule
    ProtoNet \cite{NIPS2017_cb8da676} & 35.24\tiny$\pm$0.63 & 30.72\tiny$\pm$0.62 & 30.27\tiny$\pm$0.62 & 28.52\tiny$\pm$0.58 & 28.44\tiny$\pm$0.63 & 26.80\tiny$\pm$0.47 & 27.31\tiny$\pm$0.58 \\
    RelationNet \cite{sung2018learning} & 34.86\tiny$\pm$0.63 & 28.28\tiny$\pm$0.62 & 27.59\tiny$\pm$0.56 & 27.66\tiny$\pm$0.58 & 25.86\tiny$\pm$0.60 & 25.98\tiny$\pm$0.54 & 27.83\tiny$\pm$0.63 \\
    MetaOptNet \cite{lee2019meta} & 36.77\tiny$\pm$0.65 & 33.34\tiny$\pm$0.69 & 33.28\tiny$\pm$0.65 & 28.78\tiny$\pm$0.53 & 28.70\tiny$\pm$0.64 & 29.45\tiny$\pm$0.69 & 28.36\tiny$\pm$0.64 \\
    Tian et al. \cite{tian2020rethinking} & 39.53\tiny$\pm$0.67 & 33.88\tiny$\pm$0.69 & 33.98\tiny$\pm$0.67 & 30.44\tiny$\pm$0.60 & 30.86\tiny$\pm$0.66 & 30.26\tiny$\pm$0.57 & 30.30\tiny$\pm$0.62 \\
    DeepEMD \cite{Zhang_2020_CVPR} & 41.19\tiny$\pm$0.71 & 34.27\tiny$\pm$0.72 & 35.19\tiny$\pm$0.71 & 30.92\tiny$\pm$0.62 & 31.82\tiny$\pm$0.70 & 31.05\tiny$\pm$0.59 & 31.07\tiny$\pm$0.63\\
    ProtoNet+FWT \cite{Tseng2020Cross-Domain} & 35.43\tiny$\pm$0.64 & 32.18\tiny$\pm$0.67 & 30.92\tiny$\pm$0.61 & 28.75\tiny$\pm$0.62 & 27.93\tiny$\pm$0.63 & 27.58\tiny$\pm$0.52 & 28.37\tiny$\pm$0.65 \\
    ProtoNet+ATA \cite{ijcai2021-149} & 35.67\tiny$\pm$0.66 & 31.56\tiny$\pm$0.68 & 30.40\tiny$\pm$0.62 & 27.20\tiny$\pm$0.56 & 26.61\tiny$\pm$0.62 & 27.88\tiny$\pm$0.55 & 28.48\tiny$\pm$0.65 \\
    S2M2 \cite{mangla2020charting} & 41.92\tiny$\pm$0.68 & \textbf{35.46\tiny$\pm$0.74} & 35.21\tiny$\pm$0.70 & \textbf{31.84\tiny$\pm$0.66} & \textbf{31.96\tiny$\pm$0.66} & 30.36\tiny$\pm$0.59 & 30.88\tiny$\pm$0.65 \\
    Meta-Baseline \cite{chen2021meta} & 38.88\tiny$\pm$0.67 & 33.44\tiny$\pm$0.72 & 33.73\tiny$\pm$0.68 & 30.41\tiny$\pm$0.61 & 30.43\tiny$\pm$0.67 & 30.00\tiny$\pm$0.58 &30.31\tiny$\pm$0.64 \\
    $stab\text{PA}^-$ \scriptsize(\textit{Ours}) & \textbf{43.43\tiny$\pm$0.69} & 35.16\tiny$\pm$0.72 & \textbf{35.74\tiny$\pm$0.68} & 31.16\tiny$\pm$0.66 & 30.44\tiny$\pm$0.64 & \textbf{32.09\tiny$\pm$0.62} & \textbf{31.71\tiny$\pm$0.67} \\
    \cmidrule(lr){2-8}
    DANN \cite{ganin2016domain} & - & 33.41\tiny$\pm$0.71 & 33.60\tiny$\pm$0.66 & 30.98\tiny$\pm$0.64 & 30.81\tiny$\pm$0.70 & 31.67\tiny$\pm$0.60 & 32.07\tiny$\pm$0.64 \\
    PCT \cite{tanwisuth2021a} & - & 35.53\tiny$\pm$0.73 & 35.58\tiny$\pm$.71 & 28.83\tiny$\pm$0.58 & 28.44\tiny$\pm$0.67 & 31.56\tiny$\pm$0.58 & 31.59\tiny$\pm$0.65 \\
    Mean Teacher \cite{NIPS2017_68053af2} & - & 33.24\tiny$\pm$0.70 & 33.13\tiny$\pm$0.67 & 31.34\tiny$\pm$0.62 & 30.91\tiny$\pm$0.67 & 30.98\tiny$\pm$0.60 & 31.57\tiny$\pm$0.61 \\
    FixMatch \cite{sohn2020fixmatch} & - & \textbf{36.05\tiny$\pm$0.73} & \textbf{35.83\tiny$\pm$0.76} & \textbf{33.79\tiny$\pm$0.64} & \textbf{33.20\tiny$\pm$0.74} & 31.81\tiny$\pm$0.60 & 32.32\tiny$\pm$0.66 \\
    STARTUP \cite{phoo2021selftraining} & - & 34.62\tiny$\pm$0.74 & 34.80\tiny$\pm$0.68 & 30.70\tiny$\pm$0.63 & 30.17\tiny$\pm$0.68 & \textbf{32.06\tiny$\pm$0.59} & \textbf{32.40\tiny$\pm$0.66} \\
    \cmidrule(lr){3-8}
    \textit{stab}PA \scriptsize(\textit{Ours}) & - & \textbf{38.02\tiny$\pm$0.76} & \textbf{38.09\tiny$\pm$0.82} & \textbf{35.44\tiny$\pm$0.76} & \textbf{34.74\tiny$\pm$0.76} & \textbf{34.81\tiny$\pm$0.69} & \textbf{35.18\tiny$\pm$0.72} \\

    \midrule
    & \multicolumn{7}{c}{5-way 5-shot}\\
    \cmidrule(lr){2-8}
    ProtoNet \cite{NIPS2017_cb8da676} & 49.21\tiny$\pm$0.59 & 39.74\tiny$\pm$0.64 & 38.98\tiny$\pm$0.64 & 34.81\tiny$\pm$0.59 & 35.85\tiny$\pm$0.59 & 34.56\tiny$\pm$0.58 & 36.27\tiny$\pm$0.66 \\
    RelationNet \cite{sung2018learning} & 47.02\tiny$\pm$0.57 & 33.95\tiny$\pm$0.60 & 32.78\tiny$\pm$0.59 & 33.58\tiny$\pm$0.60 & 30.15\tiny$\pm$0.55 & 30.44\tiny$\pm$0.55 & 35.42\tiny$\pm$0.70 \\
    MetaOptNet \cite{lee2019meta} & 52.00\tiny$\pm$0.59 & 43.21\tiny$\pm$0.69 & 42.97\tiny$\pm$0.63 & 36.48\tiny$\pm$0.57 & 36.56\tiny$\pm$0.65 & 36.75\tiny$\pm$0.63 & 38.48\tiny$\pm$0.68 \\
    Tian et al.\cite{tian2020rethinking} & 56.89\tiny$\pm$0.61 & 45.79\tiny$\pm$0.69 & 44.27\tiny$\pm$0.63 & 38.27\tiny$\pm$0.64 & 38.99\tiny$\pm$0.63 & 38.80\tiny$\pm$0.61 & 41.56\tiny$\pm$0.72 \\
    DeepEMD \cite{Zhang_2020_CVPR} & 58.76\tiny$\pm$0.61 & 47.47\tiny$\pm$0.71 & 45.39\tiny$\pm$0.65 & 38.87\tiny$\pm$0.63 & 40.06\tiny$\pm$0.66 & 39.20\tiny$\pm$0.58 & 41.62\tiny$\pm$0.72 \\
    ProtoNet+FWT \cite{Tseng2020Cross-Domain} & 51.40\tiny$\pm$0.61 & 41.50\tiny$\pm$0.68 & 40.32\tiny$\pm$0.60 & 36.07\tiny$\pm$0.62 & 35.80\tiny$\pm$0.60 & 34.60\tiny$\pm$0.56 & 37.36\tiny$\pm$0.67 \\
    ProtoNet+ATA \cite{ijcai2021-149} & 51.19\tiny$\pm$0.63 & 41.19\tiny$\pm$0.68 & 38.06\tiny$\pm$0.61 & 32.74\tiny$\pm$0.56 & 33.98\tiny$\pm$0.67 & 35.36\tiny$\pm$0.56 & 36.87\tiny$\pm$0.68 \\
    S2M2 \cite{mangla2020charting} & 60.82\tiny$\pm$0.58 & 47.84\tiny$\pm$0.70 & \textbf{46.32\tiny$\pm$0.67} & \textbf{40.09\tiny$\pm$0.66} & \textbf{41.63\tiny$\pm$0.64} & 40.01\tiny$\pm$0.60 & 42.68\tiny$\pm$0.67 \\
    Meta-Baseline \cite{chen2021meta} & 55.75\tiny$\pm$0.60 & 45.33\tiny$\pm$0.73 & 42.62\tiny$\pm$0.63 & 37.29\tiny$\pm$0.60 & 38.21\tiny$\pm$0.66 & 38.35\tiny$\pm$0.62 & 41.54\tiny$\pm$0.71 \\
    $stab\text{PA}^-$ \scriptsize(\textit{Ours}) & \textbf{61.87\tiny$\pm$0.57} & \textbf{48.02\tiny$\pm$0.73} & 46.27\tiny$\pm$0.67 & 38.22\tiny$\pm$0.66 & 39.88\tiny$\pm$0.63 & \textbf{41.75\tiny$\pm$0.59} & \textbf{44.09\tiny$\pm$0.69}\\
    \cmidrule(lr){2-8}
    DANN \cite{ganin2016domain} & - & 45.09\tiny$\pm$0.48 & 42.71\tiny$\pm$0.65 & 39.11\tiny$\pm$0.61 & 39.49\tiny$\pm$0.69 & 41.40\tiny$\pm$0.59 & 43.68\tiny$\pm$0.73 \\
    PCT \cite{tanwisuth2021a} & - & 48.06\tiny$\pm$0.68 & 46.25\tiny$\pm$0.64 & 34.10\tiny$\pm$0.58 & 35.59\tiny$\pm$0.66 & 40.85\tiny$\pm$0.58 & 43.30\tiny$\pm$0.74 \\
    Mean Teacher \cite{NIPS2017_68053af2} & - & 44.80\tiny$\pm$0.69 & 43.16\tiny$\pm$0.61 & 39.30\tiny$\pm$0.61 & 39.37\tiny$\pm$0.66 & 39.98\tiny$\pm$0.60 & 42.50\tiny$\pm$0.68 \\
    FixMatch \cite{sohn2020fixmatch} & - & \textbf{48.45\tiny$\pm$0.70} & \textbf{47.17\tiny$\pm$0.68} & \textbf{43.13\tiny$\pm$0.67} & \textbf{43.20\tiny$\pm$0.69} & 41.48\tiny$\pm$0.60 & 44.68\tiny$\pm$0.72 \\
    STARTUP \cite{phoo2021selftraining} & - & 47.18\tiny$\pm$0.71 & 45.00\tiny$\pm$0.64 & 38.10\tiny$\pm$0.62 & 38.84\tiny$\pm$0.70 & \textbf{41.94\tiny$\pm$0.63} & \textbf{44.71\tiny$\pm$0.73} \\
    \cmidrule(lr){3-8}
    \textit{stab}PA \scriptsize(\textit{Ours}) & - & \textbf{49.83\tiny$\pm$0.67} & \textbf{50.78\tiny$\pm$0.74} & \textbf{44.02\tiny$\pm$0.71} & \textbf{45.55\tiny$\pm$0.70} & \textbf{45.64\tiny$\pm$0.63} & \textbf{48.97\tiny$\pm$0.69} \\
    \bottomrule
    \end{tabular}
    }
    \label{tab:office-home}
\end{table*}

We compare our approach to a broad range of related methods. Methods in the first group \cite{NIPS2017_cb8da676,sung2018learning,lee2019meta,tian2020rethinking,Zhang_2020_CVPR,Tseng2020Cross-Domain,ijcai2021-149,mangla2020charting,chen2021meta} do not use the unlabeled auxiliary data during meta-training, while methods in the second group \cite{ganin2016domain,tanwisuth2021a,NIPS2017_68053af2,sohn2020fixmatch,phoo2021selftraining,islam2021dynamic} utilize the unlabeled target images to facilitate crossing the domain gap. Note that methods in the second group are only different in representation learning, and adopt the same evaluation paradigm as ours, i.e., training a linear classifier on the support set. We also implement a baseline method, termed $stab\text{PA}^-$, where we do not apply domain alignment and only train the feature extractor on augmented source images, which is also equivalent to applying strong augmentation to Tian el al. \cite{tian2020rethinking}. We set $\beta=0.5$, $\lambda=0.2$ and $m=0.1$ as default for our approach. All compared methods are implemented with the same backbone and optimizer. 
Implementation details including augmentation techniques can be found in the appendix.

The comparison results are shown in Tables \ref{tab:domainnet} and \ref{tab:office-home}. The `r-r' setting denotes all images are from the source domain, and thus is not available for methods in the second group. In Table \ref{tab:domainnet}, we can see that the performance of conventional FSL methods drops quickly when there is a domain shift between support and query sets. The proposed \textit{stab}PA leveraging unlabeled target images for domain alignment can alleviate this problem, improving the previous best FSL baseline \cite{chen2021meta} by 7.05\% across 6 CSCS-FSL situations. Similar results can be found on the Office-Home dataset in Table \ref{tab:office-home}, where the \textit{stab}PA outperforms the previous best FSL method, S2M2 \cite{mangla2020charting}, by 3.90\% on average. 
When comparing our approach with methods in the second group, we find that the \textit{stab}PA outperforms them in all situations, improving 5-shot accuracy by 5.98\% over the previous best method FixMatch \cite{sohn2020fixmatch} on DomainNet. These improvements indicate that the proposed bi-directional prototypical alignment is an effective approach to leveraging unlabeled images to reduce domain gap for CDCS-FSL. 

\subsection{Analysis}

\subsubsection{Has \textit{stab}PA learned compact and aligned representations?}
\begin{figure*}[t]
    \centering
    \includegraphics[width=\linewidth]{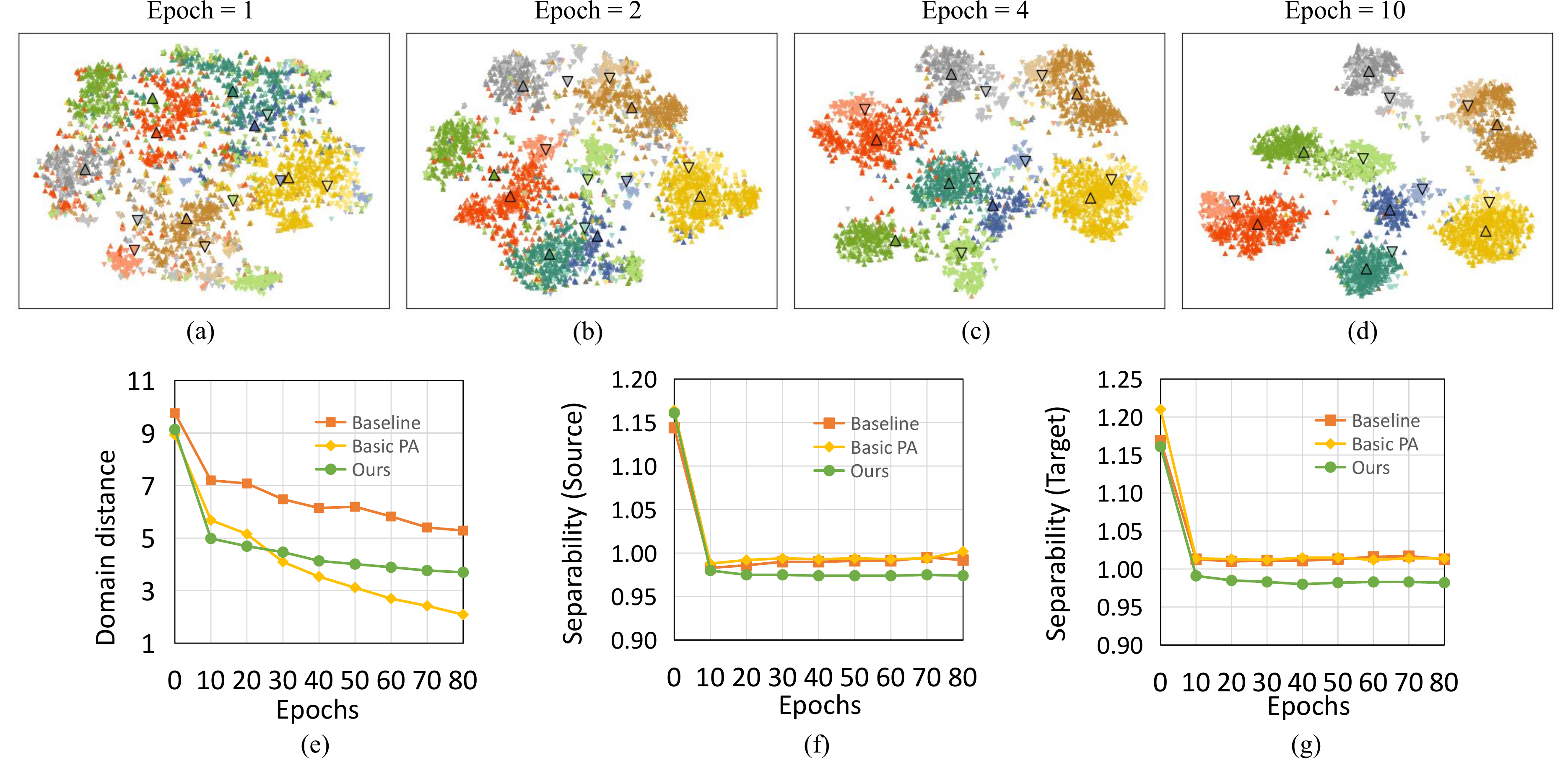}
    \caption{(a)-(d) t-SNE visualization of feature distribution at different training epochs. Samples of the same class are painted in similar colors, where darker triangles represent source samples and lighter reverted triangles represent target samples (best viewed in color). Class centers are marked in black border. (e) Domain distance on novel classes. (f)-(g) Separability among novel classes in the source and target domains. Separability is represented by the average distance ratio, the lower the better.}
    \label{fig:vis}
\end{figure*}

To verify whether \textit{stab}PA indeed learns compact and aligned representations, we visualize the feature distributions through the meta-training process using t-SNE \cite{van2008visualizing}. From Figure \ref{fig:vis} (a)-(d), it can be seen that in the beginning, samples from different classes are heavily mixed. There are no distinct classification boundaries between classes. Besides, samples from two domains are far away from each other, indicating the existence of a considerable domain shift (such as the classes in green and orange). However, as training continues, samples from the same class begin to aggregate together, and the margins between different classes are increasing. In other words, compact feature representation can be obtained by the \textit{stab}PA. Moreover, we can see that samples from different domains are grouped into their ground-truth classes, even though no label information is given for the target domain data. These observations demonstrate that \textit{stab}PA is indeed capable to learn compact and aligned representations.

\subsubsection{Can \textit{stab}PA learn generalizable representations for novel classes?}
To validate the generalization capability of the representations learned by \textit{stab}PA, we propose two quantitative metrics, Prototype Distance (PD) and Average Distance Ratio (ADR), which indicate the domain distance and class separability among novel classes, respectively. A small PD value means the two domains are well aligned to each other, and a ADR less than 1 indicates most samples are classified into their ground-truth classes. Detailed definitions about these two metrics can be found in the appendix.



We compare \textit{stab}PA with a FSL baseline \cite{tian2020rethinking} that does not leverage target images, and the BasicPA which aligns two domains by simply minimizing the point-to-point distance between prototypes in two domains \cite{xie2018learning}. The results are presented in Figure \ref{fig:vis} (e)-(g). It is noticed that all these methods can achieve lower domain distance as training processes, and BasicPA gets the lowest domain distance at the end. However, BasicPA does not improve the class separability as much as our approach, as shown in Figure \ref{fig:vis} (f)-(g). The inferior class separability can be understood that BasicPA merely aims to reduce the feature distance between two domains, without taking account of the intra-class variance and inter-class distances in each domain. Instead of aligning centers adopted by BasicPA, the proposed \textit{stab}PA considers the feature-to-prototype distances across different domains and classes, so the domain alignment and class separability can be improved at the same time.


\subsubsection{Number of unlabeled target data.}
\label{sec:data_number}

To test the robustness to the number of unlabeled samples, we gradually drop data from the auxiliary set in two ways: (i) randomly drop samples from the auxiliary set, (ii) select a subset of base classes and then manually remove samples that are from the selected classes. Table \ref{tab:ratio} shows the average accuracy of $stab$PA on DomainNet over 6 situations. Unsurprisingly, decreasing the number of samples will lead to performance drop (about 2.4 points from 100\% to 10\%). However, with only 10\% samples remained, our approach still outperforms FixMatch which uses 100\% auxiliary data. We can also see that removing whole classes leads to more performance drop than randomly removing samples, probably due to the class imbalance problem caused by the former. Nevertheless, the difference is very small (about 0.3 points), indicating that our approach is robust to the number of base classes.

\subsubsection{Pseudo label accuracy.} In Table \ref{label_acc}, we show the pseudo label accuracies of the target domain images obtained by the fixed classifier and the online classifier during the training process. We can see that the fixed classifier is better than the online classifier at the early training epochs. However, as the training goes on, the online classier gets more accurate and outperforms the fixed classifier. This is because the online classifier is updated along the representation alignment process and gradually fits the data distribution of the target domain. After training with 50 epochs, the online classifier achieves 53.9\% top-5 accuracy. To further improve the reliability of pseudo labels, we set a threshold to filter out pseudo labels with low confidence. Therefore, the actual pseudo label accuracy is higher than 53.9\%.

\begin{table}[t]
    \centering
    \small
    \caption{The influence of the number of unlabeled samples and the number of base classes in the auxiliary set. We report average accuracy on DomainNet over 6 situations.}
    \resizebox{.95\linewidth}{!}{
    \begin{tabular}{p{1.2cm}<{\centering}c|p{1cm}<{\centering}p{1cm}<{\centering}p{1cm}<{\centering}p{1cm}<{\centering}|p{1cm}<{\centering}p{1cm}<{\centering}p{1cm}<{\centering}p{1cm}<{\centering}p{1cm}<{\centering}}
    \toprule
    & \multicolumn{1}{c}{} & \multicolumn{4}{c}{number of samples} & \multicolumn{5}{c}{number of base classes}  \\
    \cmidrule(lr){3-6} \cmidrule(lr){7-11}
    & FixMatch \cite{sohn2020fixmatch} & 10\% & 40\% & 70\% & 100\% & 0\% & 10\% & 40\% & 70\% & 100\% \\
    \midrule
    1-shot & 47.72 & 51.76 & 52.97 & 53.42 & 53.92 & 50.74 & 51.59 & 52.48 & 53.24 & 53.92 \\
    5-shot & 62.58 & 65.96 & 67.56 & 67.96 & 68.55 & 65.04 & 65.68 & 67.07 & 67.87 & 68.55 \\
    \bottomrule
    \end{tabular}
    }
    \label{tab:ratio}
\end{table}

\begin{table}[t]
    \centering
    \caption{Pseudo label accuracy on DomainNet real-painting.}
    \resizebox{.65\linewidth}{!}{
    \begin{tabular}{cp{1cm}<{\centering}cp{1cm}<{\centering}p{1cm}<{\centering}p{1cm}<{\centering}p{1cm}<{\centering}p{1cm}<{\centering}}
    \toprule
    & fixed & epoch=0 & 10 & 20 & 30 & 40 & 50 \\
    \midrule
    Top-1 & 23.5 & 4.9 & 24.2 & 30.8 & 34.4 & 35.9 & 37.2 \\
    Top-5 & 40.0 & 14.4 & 41.9 & 48.2 & 51.4 & 52.8 & 53.9 \\
    \bottomrule
    \end{tabular}
    }
    \label{label_acc}
\end{table}

\subsubsection{Ablation studies.}
\label{ablation}


We conduct ablation studies on key components of the proposed \textit{stab}PA. The results on DomainNet are shown in Table \ref{tab:ablation}. As all key components are removed (the first row), our approach is similar to Tian et al. \cite{tian2020rethinking} that trains feature extractor with only the source data. When the unlabeled target data are available, applying either source-to-target alignment or target-to-source alignment can improve the performance evidently. Interestingly, we can see that the target-to-source alignment is more effective than the source-to-target alignment (about 1.2 points on average). This is probably because the source prototypes estimated by the ground truth labels are more accurate than the target prototypes estimated by the pseudo labels. Improving the quality of target prototypes may reduce this gap. Combing these two alignments together, we can get better results, indicating that the two alignments are complementary to each other. Finally, the best results are obtained by combining the strong data augmentation techniques, verifying that strong data augmentation can further strengthen the cross-domain alignment.


\begin{table}[t]
    \small
    \centering
    \caption{Ablation studies on DomainNet with 95\%  confidence interval.}
    \resizebox{.75\linewidth}{!}{
    \begin{tabular}{p{1cm}<{\centering}p{1cm}<{\centering}p{1cm}<{\centering}cc|cc}
    \toprule
    & & & \multicolumn{2}{c}{real-sketch} & \multicolumn{2}{c}{sketch-real}\\
    \cmidrule(lr){4-5} \cmidrule(lr){6-7}
    $\ell_{s-t}$  & $\ell_{t-s}$ & aug & 1-shot & 5-shot & 1-shot & 5-shot\\
    \midrule
    $\times$ & $\times$ & $\times$ & 40.23\tiny$\pm$0.73 & 50.41\tiny$\pm$0.80 & 41.90\tiny$\pm$0.86 & 56.95\tiny$\pm$0.84\\
    $\times$ & $\times$ & \checkmark & 41.74\tiny$\pm$0.78 & 51.03\tiny$\pm$0.85 & 42.17\tiny$\pm$0.95 & 57.11\tiny$\pm$0.93\\
    \checkmark & $\times$ & $\times$ & 42.86\tiny$\pm$0.78 & 52.16\tiny$\pm$0.78 & 44.83\tiny$\pm$0.95 & 60.87\tiny$\pm$0.91\\
    $\times$ & \checkmark & $\times$ & 44.20\tiny$\pm$0.77 & 54.83\tiny$\pm$0.79 & 44.45\tiny$\pm$0.92 & 61.97\tiny$\pm$0.90\\
    \checkmark & \checkmark & $\times$ & 47.01\tiny$\pm$0.84 & 56.68\tiny$\pm$0.81 & 47.59\tiny$\pm$1.00 & 64.32\tiny$\pm$0.86\\
    \checkmark & \checkmark & \checkmark & \textbf{50.85\tiny$\pm$0.86} & \textbf{61.37\tiny$\pm$0.82} & \textbf{51.71\tiny$\pm$1.01} & \textbf{68.93\tiny$\pm$0.87}\\
    \bottomrule
    \end{tabular}
    }
    \label{tab:ablation}
\end{table}

\section{Conclusions}

In this work, we have investigated a new problem in FSL, namely CDCS-FSL, where a domain gap exists between the support set and query set. To tackle this problem, we have proposed \textit{stab}PA, a prototype-based domain alignment framework to learn compact and cross-domain aligned representations. On two widely-used multi-domain datasets, we have compared our approach to multiple elaborated baselines. Extensive experimental results have demonstrated the advantages of our approach. Through more in-depth analysis, we have validated the generalization capability of the representations learned by \textit{stab}PA and the effectiveness of each component of the proposed model.

\section*{Acknowledgements}
This work was supported in part by the National Natural Science Foundation of China under Grants 61721004, 61976214, 62076078, 62176246 and in part by the CAS-AIR.

%
%
\bibliographystyle{splncs04}
\bibliography{egbib}



\appendix

\section*{Appendix 1: Details of our approach}
\label{details}

\paragraph{Hyper-parameters.} In our implementation, ResNet-18 \cite{he2016deep} is adopted as the backbone, which outputs a 512-d feature vector. Before feeding the vector for prototypical alignment, we apply $\ell_2$ normalization for the feature vector and prototypes. The temperature $\tau$ for $\ell_{s-t}$ and $\ell_{t-s}$ is 0.25 and 0.1, respectively. The max training steps $T_{max}$ is set as 50,000 for the DomainNet and 1,000 for the Office-Home, which are roughly equal to $training\ epochs \times dataset\ size / batch\ size$. The confidence threshold $\beta$ for $\ell_{t-s}$ is set as 0.5. $\lambda$ is equal to 0.2 to balance the pseudo labels generated by the initial classifier and the online updated classifier. The momentum term $m$ is set as 0.1. These hyper-parameters are tuned based on performance on the validation set.

\paragraph{Training.} We train our approach for 50 epochs on the DomainNet dataset. On the smaller Office-Home dataset, we train the model for 100 epochs. Adam \cite{KingmaB14} is adopted as the default optimizer with the learning rate as 1e-3. The batch size is set as 256, where source data and target data have the same number in a batch (128). 

\paragraph{Evaluation.} During evaluation, we fix the feature extractor and apply $\ell_2$ normalization to the output feature vector. The linear classification head for each few-shot task (episode) is randomly initialized, and trained on the support features for 1000 steps with logistic regression. 15 query samples per class are used to evaluate the performance of the learned classifier. We finally report the average 5-way 1-shot and 5-way 5-shot accuracies over 600 episodes with 95\% confidence intervals.

\begin{table}[]
    \centering
    \caption{The impact of interpolation coefficient $\lambda$.}
    \begin{tabular}{p{1.2cm}<{\centering}|p{1.7cm}<{\centering}p{1.7cm}<{\centering}|p{1.7cm}<{\centering}p{1.7cm}<{\centering}}
    \toprule
    & \multicolumn{2}{c}{painting-real} & \multicolumn{2}{c}{real-painting} \\
    \cmidrule(lr){2-3} \cmidrule(lr){4-5}
    $\lambda$ & 1-shot & 5-shot & 1-shot & 5-shot \\
    \midrule
    0.0 & 53.91\tiny$\pm$1.03 & 72.64\tiny$\pm$0.85 & 53.73\tiny$\pm$0.90 & 64.95\tiny$\pm$0.79\\
    0.2 & 54.44\tiny$\pm$1.00 & \textbf{73.63\tiny$\pm$0.82} & 53.86\tiny$\pm$0.89 & \textbf{65.65\tiny$\pm$0.74}\\
    0.4 & \textbf{54.55\tiny$\pm$1.03} & 73.50\tiny$\pm$0.83 & \textbf{53.99\tiny$\pm$0.90} & 64.87\tiny$\pm$0.78\\
    0.6 & 50.50\tiny$\pm$1.03 & 69.11\tiny$\pm$0.89 & 50.47\tiny$\pm$0.87 & 61.26\tiny$\pm$0.81\\
    0.8 & 50.07\tiny$\pm$1.00 & 68.50\tiny$\pm$0.90 & 50.40\tiny$\pm$0.87 & 60.52\tiny$\pm$0.80\\
    1.0 & 49.79\tiny$\pm$1.03 & 68.42\tiny$\pm$0.90 & 50.28\tiny$\pm$0.89 & 60.60\tiny$\pm$0.79\\
    \bottomrule
    \end{tabular}
    \label{tab:lambda}
\end{table}

\section*{Appendix 2: Updating pseudo label}

Since we resort to pseudo labels for prototype estimation and feature alignment, ensuring the pseudo label accuracy is very important to the effectiveness of our bi-directional prototypical alignment strategy. Pseudo labels can be predicted with a fixed classifier pre-trained on the source base dataset, as in \cite{phoo2021selftraining}, or a classifier that is online updated along the representation learning. In our implementation, we combine them together by linearly interpolating their pseudo labels. We assess the effectiveness of this combination strategy by changing the interpolation coefficient $\lambda$ from zero to one. When the interpolation coefficient $\lambda=0\,(\text{or} 1)$, our approach degenerates to only using the fixed (or online updated) classifier. The results on the DomainNet are shown in Table \ref{tab:lambda}. 

It can be noticed that the performance grows as we increase $\lambda$ from zero and the best performance can be achieved when $\lambda \in [0.2, 0.4]$. The improvement demonstrates that updating the fixed pseudo labels with an online classifier is useful to get better pseudo labels. However, when $\lambda$ gets too large, the performance drops very quickly, which means we can not only depend on the online classifier. The possible reason is that the pseudo labels predicted by the online classifier change rapidly, and thus impose adverse impacts on the training stability.

\section*{Appendix 3: Hyper-parameter sensitivity}

\begin{figure}[t]
    \centering
    \includegraphics[width=.8\linewidth]{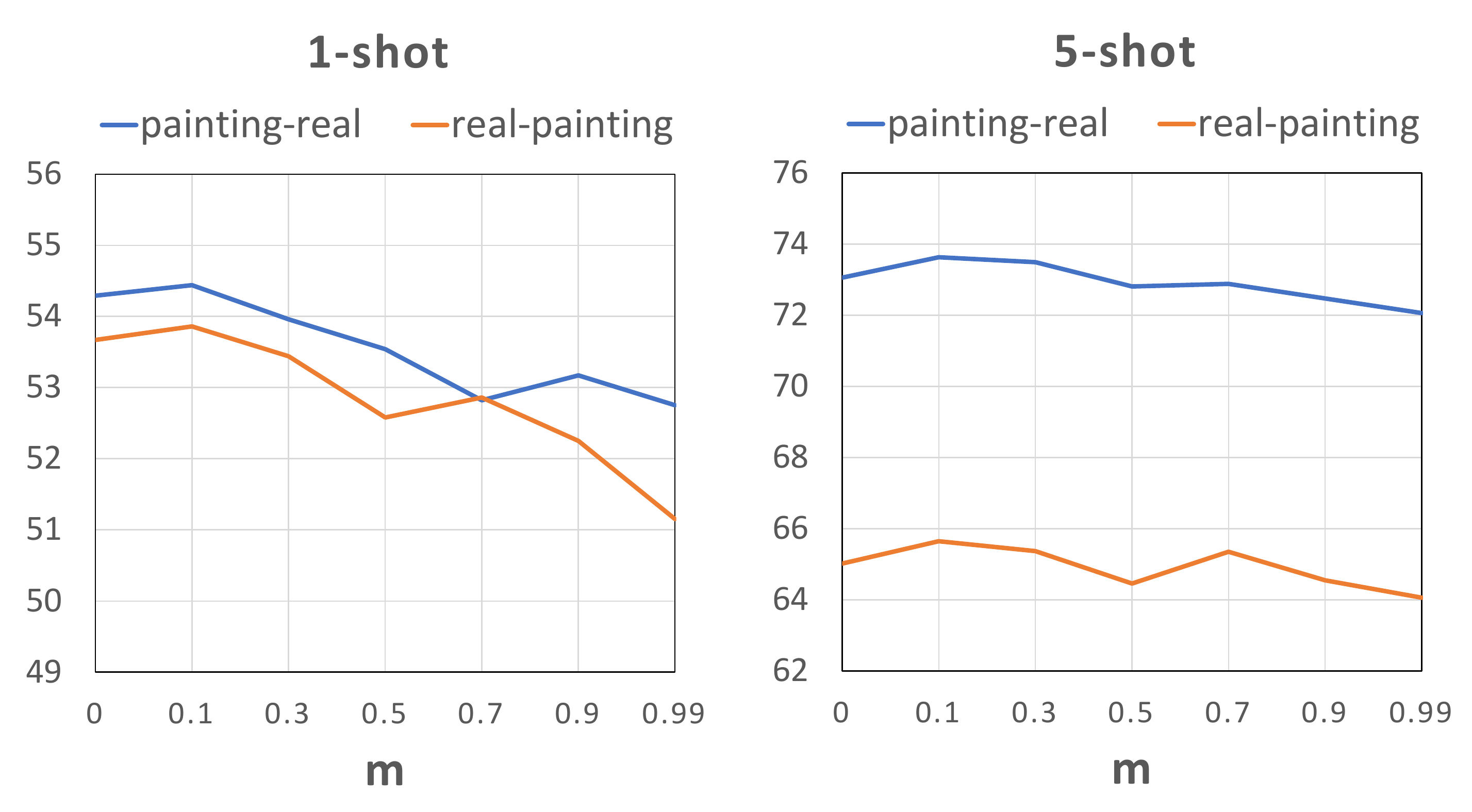}
    \caption{The sensitivity of momentum coefficient $m$.}
    \label{fig:m}
\end{figure}

\begin{figure}[t]
    \centering
    \includegraphics[width=.8\linewidth]{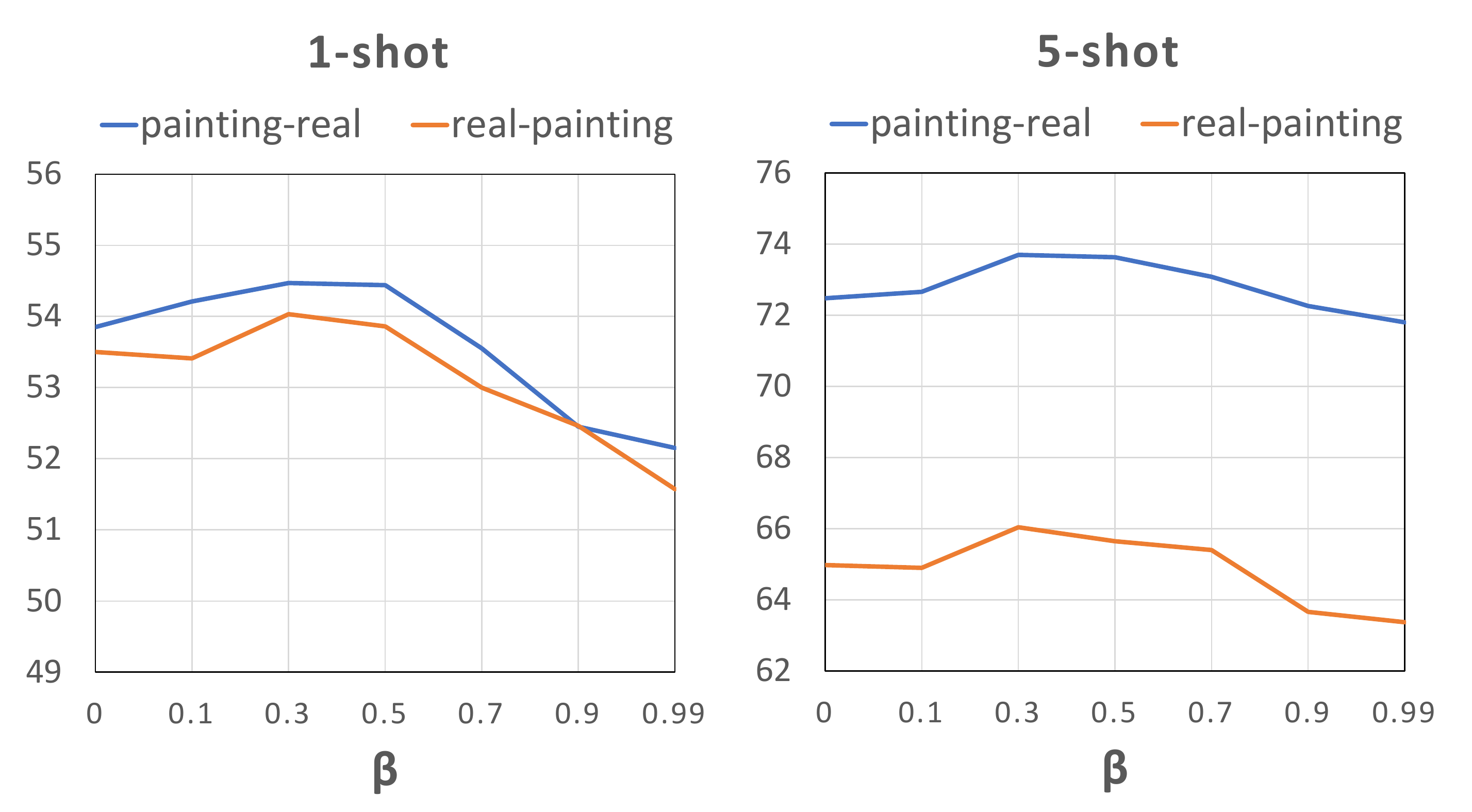}
    \caption{The sensitivity of confidence threshold $\beta$.}
    \label{fig:threshold}
\end{figure}

To analyse the sensitivity of a hyper-parameter, we change its value from the minimum to the maximum and keep other hyper-parameters unchanged. We test the performance of each value on the DomainNet \textit{real-painting} and \textit{painting-real}. The experimental results are shown in Figures \ref{fig:m} and \ref{fig:threshold}. For the momentum coefficient $m$, a small $m$ is usually better than a large one. The gap between the best performance ($m=0.1$) and the worst performance ($m=0.99$) is 2.2 points in 1-shot and 1.6 points in 5-shot. For the confidence threshold $\beta$, the performance grows in the range of $[0, 0.3]$ and decreases rapidly in the rage of $[0.5, 0.99]$. The difference between the best and the worst results are 2.4 points in 1-shot and 2.3 points in 5-shot, which are a little bit larger than the differences of $m$. However, the performance of the proposed approach is still competitive even with the worst hyper-parameters, indicating that our approach is not very sensitive to hyper-parameters.

\section*{Appendix 4: Prototype Distance (PD) and Average Distance Ratio (ADR)}
To measure domain distance, we first calculate prototypes $p^s_k$ and $p^t_k$ for each novel class in the source and target domains. Then we obtain the Euclidean distance between the two prototypes per class and compute the average distance over all novel classes. We refer to this metric as Prototype Distance (PD), which can be formulated as:
\begin{equation}
    PD = \frac{1}{|\mathcal{Y_N}|} \sum_{k\in\mathcal{Y_N}} ||p^s_k-p^t_k||,
\end{equation} 
where $\mathcal{Y_N}$ is the set of novel classes. A small PD value means the two domains are well aligned to each other.

To represent class separability, for each sample $(x_i, y_i)$, we calculate the ratio between its distance to the prototype $p_{y_i}$, and the distance to the closest neighbouring prototype. Then an average is computed over all samples in novel classes, which is termed Average Distance Ratio (ADR). Formally,
\begin{equation}
    ADR = \frac{1}{|\mathcal{X_N}|} \sum_{x_i\in \mathcal{X_N}} \frac{||f(x_i) - p_{y_i}||}{\min_{k\neq y_i} ||f(x_i)-p_k||},
\end{equation}
where $\mathcal{X_N}$ is the set of samples of novel classes. When ADR is less than 1, most samples are classified into their ground-truth classes. We calculate ADR for two domains separately to validate whether the learned features can generalize in each domain.

\section*{Appendix 5: Baselines}

For a fair comparison, we implement all the baseline methods with the same ResNet-18 backbone adopted in our approach. But the augmentation strategies may be different for different methods, as some methods \cite{Zhang_2020_CVPR,mangla2020charting,sohn2020fixmatch,phoo2021selftraining,islam2021dynamic} have specified particular augmentation in their papers, where FixMatch\cite{sohn2020fixmatch} adopt the same augmentation techniques as ours. When no augmentation is specified, we simply apply CenterCrop and Normalization to the input images.

\paragraph{ProtoNet and RelationNet.} ProtoNet\cite{NIPS2017_cb8da676} and RelationNet\cite{sung2018learning} are two representative meta-learning methods, which are trained on a series of few-shot tasks (episodes). We implement these two methods based on publicly-available codes \footnote{https://github.com/wyharveychen/CloserLookFewShot}. During training, we randomly sample episodes from the base set, each of which contains $N=5$ classes and $K=5$ samples per class serving as the support set, and another 15 samples per class as the query set. We also train ProtoNet and RelationNet for 50 epochs on the DomainNet dataset and 100 epochs on the Office-Home dataset. The number of training episodes of each epoch is particularly defined to make sure the number of seen samples (both the support and query samples) in an epoch is roughly equal to the size of the dataset.

\paragraph{MetaOptNet.} MetaOptNet\cite{lee2019meta} aims to learn an embedding function that generalizes well to novel categories with closed-form linear classifiers (e.g., SVMs). We implement this method based on the official code \footnote{https://github.com/kjunelee/MetaOptNet} but replace the backbone network and optimizer to be the same as our approach. Similar to ProtoNet and RelationNet, the training process of MetaOptNet is also episodic.

\paragraph{Tian et al.} Tian et al.\cite{tian2020rethinking} follows the transfer learning paradigm, which trains a base model by classifying base classes, and then leverages the learned representations to classify novel classes by learning a new classification head. We train this baseline with the same optimization method as our approach except that the batch size is set as 128 as only source data are used for training.

\paragraph{DeepEMD.} DeepEMD\cite{Zhang_2020_CVPR} is also a meta-learning method, which aims to compute the query-support similarity based on Earth Mover’s Distance (EMD). It contains two training phases: (i) pre-training the feature extractor by classifying base classes (similar to Tian et al.) and (ii) meta-training the whole model on training episodes. We use the output model of Tian et al. as the pre-trained model and then follow the official implementation \footnote{https://github.com/icoz69/DeepEMD} to finetune the model via meta-training.

\paragraph{FWT and ATA.} FWT\cite{Tseng2020Cross-Domain} and ATA\cite{ijcai2021-149} are two CD-FSL methods, which aims to learn generalized representations during meta-training so that the model can generalize to a new domain. To this end, FWT proposes a feature-wise transformation layer, of which the parameters can be manually set, or learned from multiple data sources. In our experiments, we choose to manually set the parameters as only data from one domain (the source domain) are labeled. ATA proposes to augment the task distributions by maximizing the training loss and meanwhile learn robust inductive bias from augmented task distributions. It does not need to access extra data sources, and thus can be trained on the base set. We implement these two methods based on their official codes\footnote{https://github.com/hytseng0509/CrossDomainFewShot, https://github.com/Haoqing-Wang/ CDFSL-ATA}, except that we train them from scratch as we find that additional pre-training will reduce performance.

\paragraph{S2M2.} S2M2\cite{mangla2020charting} follows the transfer learning paradigm, which leverages the data augmentation technique, MixUp\cite{verma2019manifold}, and self-supervised learning tasks (e.g., rotation) to learn generalized representation for few-shot learning. We follow the same augmentation and implementation as the official codes\footnote{https://github.com/nupurkmr9/S2M2\_fewshot}.

\paragraph{DANN.} We use a three-layer fully connected network as the domain discriminator to implement DANN, following the Pytorch implementation \footnote{https://github.com/thuml/CDAN} released by \cite{long2018conditional}. The gradient reverse layer \cite{ganin2016domain} is adopted to train the feature vector and domain discriminator in an adversarial manner. To stabilize training, the weight of the adversarial loss starts from zero, and gradually grows to one.

\paragraph{PCT.} PCT\cite{tanwisuth2021a} is a generic domain adaptation method that can deal with single-source, multi-source, class-imbalance and source-private domain adaptation problems. Similar to our approach, PCT also aligns features via prototypes. However, it only aligns features from the target domain to the prototypes trained with labeled source domain data. We implement this baseline according to the official codes\footnote{https://github.com/korawat-tanwisuth/Proto\_DA}.

\paragraph{Mean Teacher, Fixmatch and STARTUP.} All of these approaches use pseudo-labeled samples to train the model. Differently, Mean Teacher\cite{NIPS2017_68053af2} predicts pseudo labels with a teacher network that is the ensemble of historical models by aggregating their model weights with exponential moving average (EMA). In our implementation, the smoothing coefficient for EMA is set as 0.99. Fixmatch\cite{sohn2020fixmatch} trains the model with a consistency loss, i.e., enforcing the network prediction for a strongly augmented sample to be consistent with the prediction of its weakly augmented counterpart. We implement Fixmatch based on a publicly available implementation\footnote{https://github.com/kekmodel/FixMatch-pytorch}. STARTUP\cite{phoo2021selftraining} adopts fixed pseudo labels that are predicted by a classifier pre-trained on the base set, and imposes a self-supervised loss on the target data. In our re-implementation, we do not utilize the self-supervised loss item since we find that it does not improve performance.

\section*{Appendix 6: Dataset partition details}
\label{partition}

\paragraph{DomainNet.} 
DomainNet contains 345 classes in total. We discard 19 classes with too few images and randomly split the rest 326 classes into three sets: 228 classes for the base set, 33 classes for the validation set, and 65 classes for the novel set. The detailed classes of each set are listed below:
\begin{equation*}
    \mathcal{Y}_{base} = 
\end{equation*}
{\fontsize{8pt}{\baselineskip}\selectfont\{aircraft carrier, airplane, alarm clock, ambulance, animal migration, ant, asparagus, axe, backpack, bat, bathtub, beach, bear, beard, bee, belt, bench, bicycle, binoculars, bird, book, boomerang, bottlecap, bowtie, bracelet, brain, bread, bridge, broccoli, broom, bus, butterfly, cactus, cake, calculator, camera, candle, cannon, canoe, car, cat, ceiling fan, cell phone, cello, chair, church, circle, clock, cloud, coffee cup, computer, couch, cow, crab, crayon, crocodile, cruise ship, diamond, dishwasher, diving board, donut, dragon, dresser, drill, drums, duck, ear, elbow, elephant, envelope, eraser, eye, fan, feather, fence, finger, fire hydrant, fireplace, firetruck, flamingo, flashlight, flip flops, flower, flying saucer, foot, fork, frog, frying pan, giraffe, goatee, grapes, grass, guitar, hamburger, hammer, hand, harp, headphones, hedgehog, helicopter, helmet, hockey puck, hockey stick, horse, hot air balloon, hot tub, hourglass, hurricane, jacket, key, keyboard, knee, ladder, lantern, laptop, leaf, leg, light bulb, lighter, lightning, lion, lobster, lollipop, mailbox, marker, matches, megaphone, mermaid, microphone, microwave, moon, motorbike, moustache, nail, necklace, nose, octagon, oven, paint can, paintbrush, palm tree, panda, pants, paper clip, parachute, parrot, passport, peanut, pear, peas, pencil, penguin, pickup truck, picture frame, pizza, pliers, police car, pond, popsicle, postcard, potato, power outlet, purse, rabbit, radio, rain, rainbow, rake, remote control, rhinoceros, rifle, sailboat, school bus, scorpion, screwdriver, see saw, shoe, shorts, skateboard, skyscraper, smiley face, snail, snake, snorkel, soccer ball, sock, stairs, stereo, stethoscope, stitches, stove, strawberry, submarine, sweater, swing set, sword, t-shirt, table, teapot, teddy-bear, television, tent, the Eiffel Tower, the Mona Lisa, toaster, toe, toilet, tooth, toothbrush, tornado, tractor, train, tree, triangle, trombone, truck, underwear, van, vase, violin, washing machine, watermelon, waterslide, whale, wheel, windmill, wine bottle, zigzag\}}
\begin{equation*}
    \mathcal{Y}_{validation} = 
\end{equation*}
{\fontsize{8pt}{\baselineskip}\selectfont\{arm, birthday cake, blackberry, bulldozer, campfire, chandelier, cooler, cup, dumbbell, hexagon, hospital, house plant, ice cream, jail, lighthouse, lipstick, mushroom, octopus, raccoon, roller coaster, sandwich, saxophone, scissors, skull, speedboat, spreadsheet, suitcase, swan, telephone, traffic light, trumpet, wine glass, wristwatch\}}
\begin{equation*}
    \mathcal{Y}_{novel} = 
\end{equation*}
{\fontsize{8pt}{\baselineskip}\selectfont\{anvil, banana, bandage, barn, basket, basketball, bed, blueberry, bucket, camel, carrot, castle, clarinet, compass, cookie, dog, dolphin, door, eyeglasses, face, fish, floor lamp, garden, garden hose, golf club, hat, hot dog, house, kangaroo, knife, map, monkey, mosquito, mountain, mouth, mug, ocean, onion, owl, piano, pig, pillow, pineapple, pool, river, rollerskates, sea turtle, sheep, shovel, sink, sleeping bag, spider, spoon, squirrel, steak, streetlight, string bean, syringe, tennis racquet, the Great Wall of China, tiger, toothpaste, umbrella, yoga, zebra\}}

\paragraph{Office-Home.}
There are 65 classes in the Office-Home dataset. We select 40 classes as the base set, 10 classes as the validation set, and 15 classes as the novel set, which are listed below:
\begin{equation*}
    \mathcal{Y}_{base} = 
\end{equation*}
{\fontsize{8pt}{\baselineskip}\selectfont\{alarm clock, bike, bottle, bucket, calculator, calendar, chair, clipboards, curtains, desk lamp, eraser, exit sign, fan, file cabinet, folder, glasses, hammer, kettle, keyboard, lamp shade, laptop, monitor, mouse, mug, paper clip, pen, pencil, postit notes, printer, radio, refrigerator, scissors, sneakers, speaker, spoon, table, telephone, toothbrush, toys, tv\}}
\begin{equation*}
    \mathcal{Y}_{validation} = 
\end{equation*}
{\fontsize{8pt}{\baselineskip}\selectfont\{bed, computer, couch, flowers, marker, mop, notebook, pan, shelf, soda\}}
\begin{equation*}
    \mathcal{Y}_{novel} = 
\end{equation*}
{\fontsize{8pt}{\baselineskip}\selectfont\{backpack, batteries, candles, drill, flipflops, fork, helmet, knives, oven, push pin, ruler, screwdriver, sink, trash can, webcam\}}

\end{document}